\documentclass{article}
\usepackage{ijcai24}

\usepackage{times}
\usepackage{soul}
\usepackage{url}
\usepackage[hidelinks]{hyperref}
\usepackage[utf8]{inputenc}
\usepackage[small]{caption}
\usepackage{graphicx}
\usepackage{amsmath}
\usepackage{amsthm}
\usepackage{booktabs}
\usepackage{algorithm}
\usepackage{algorithmic}
\usepackage[switch]{lineno}

\usepackage{booktabs}
\usepackage{multirow}
\usepackage{bbding} 

\usepackage[nameinlink]{cleveref}
\usepackage{hyperref}
\crefname{figure}{Figure}{Figures}
\crefname{example}{Example}{Example}
\crefname{theorem}{Theorem}{Theorem}
\crefname{corollary}{Corollary}{Corollary}
\crefname{lemma}{Lemma}{Lemma}
\crefname{proposition}{Proposition}{Proposition}
\crefname{assumption}{Assumption}{Assumption}
\crefname{section}{Section}{Section}
\crefname{algorithm}{Algorithm}{Algorithm}

\usepackage{enumitem}
\newlist{propenum}{enumerate}{1} 
\setlist[propenum]{label=\alph*{\rm)}, ref=\theproposition(\alph*)}
\crefalias{propenumi}{proposition}
\newlist{corenum}{enumerate}{1} 
\setlist[corenum]{label=\alph*{\rm)}, ref=\thecorollary(\alph*)}
\crefalias{corenumi}{corollary}
\newlist{lemenum}{enumerate}{1} 
\setlist[lemenum]{label=\alph*{\rm)}, ref=\thelemma(\alph*)}
\crefalias{lemenumi}{lemma}

\usepackage{amsthm,thmtools}

\declaretheorem[name=Theorem]{theorem}
\declaretheorem[name=Definition,style=definition]{definition}

\usepackage{amsfonts}
\usepackage{todonotes}

\title{Decision-focused Graph Neural Networks for Combinatorial Optimization\thanks{Preprint.}}

\author{
Yang Liu$^1$\and
Chuan Zhou$^1$\and
Peng Zhang$^2$\and
Shirui Pan$^3$\and
Zhao Li$^4$\And
Hongyang Chen$^5$
\affiliations
$^1$Academy of Mathematics and Systems Science, Chinese Academy of Sciences\\
$^2$Guangzhou University
$^3$Griffith University
$^4$Hangzhou Yugu Technology
$^5$Zhejiang Lab
\emails
\{liuyang2020, zhouchuan\}@amss.ac.cn, p.zhang@gzhu.edu.cn, s.pan@griffith.edu.au, lzjoey@gmail.com, dr.h.chen@ieee.org
}

\begin{document}

\maketitle

\begin{abstract}

    In recent years, there has been notable interest in investigating combinatorial optimization (CO) problems by neural-based framework. An emerging strategy to tackle these challenging problems involves the adoption of graph neural networks (GNNs) as an alternative to traditional algorithms, a subject that has attracted considerable attention. Despite the growing popularity of GNNs and traditional algorithm solvers in the realm of CO, there is limited research on their integrated use and the correlation between them within an end-to-end framework. The primary focus of our work is to formulate a more efficient and precise framework for CO by employing decision-focused learning on graphs. Additionally, we introduce a decision-focused framework that utilizes GNNs to address CO problems with auxiliary support. To realize an end-to-end approach, we have designed two cascaded modules: (a) an unsupervised trained graph predictive model, and (b) a solver for quadratic binary unconstrained optimization. Empirical evaluations are conducted on various classical tasks, including maximum cut, maximum independent set, and minimum vertex cover. The experimental results on classical CO problems (i.e. MaxCut, MIS, and MVC) demonstrate the superiority of our method over both the standalone GNN approach and classical methods.
\end{abstract}

\section{Introduction}\label{section:Introduction}



There is a growing interest that has emerged at the intersection of operations research and deep learning to address combinatorial optimization problems (COPs) recently. Demonstrations of successful COP applications have spanned diverse domains, including transportation \cite{wang2021deep,bi2022combinatorial}, healthcare \cite{juan2015review,lee2021mind}, and social analysis \cite{li2018combinatorial,du2022introduction}, providing valuable insights for future advancements. The problem set encompasses several pivotal tasks, such as maximum cut \cite{hadlock1975finding}, maximum independent set \cite{tarjan1977finding}, and minimum vertex cover \cite{cai2013numvc}.



Graph neural networks (GNNs) have attracted significant attention in the realm of deep learning, owing to their successful execution in diverse graph-based tasks, such as node classification, link prediction, and graph classification. Furthermore, researchers have employed GNNs to address combinatorial optimization problems, exemplified by PI-GNN \cite{schuetz2022combinatorial}, an innovative approach inspired by physics. PI-GNN designs a GNN architecture capable of accurately solving COPs with provable guarantees. Within PI-GNN, a graph neural network is introduced, utilizing a Hamiltonian to encode COPs and subsequently solving them using a simulated annealing algorithm.



As mentioned earlier, GNNs have recently emerged as a potent tool for tackling COPs. However, current GNN-based methods often face limitations due to their reliance on heuristics and the absence of theoretical guarantees on performance. While some researchers have attempted to solve COPs exclusively using GNNs, significant debate, and uncertainty persist regarding the efficacy of GNNs compared to traditional combinatorial optimization algorithms \cite{angelini2022cracking,boettcher2022inability}. For instance, Angelini \cite{angelini2022cracking} has asserted that a simple greedy algorithm might outperform GNNs. The authors underscore the importance of comprehending the conditions under which GNNs can effectively tackle complex problems and whether fundamental limitations exist in their capabilities. In this paper, we present our distinctive perspective on addressing classical COPs.




In our research, our emphasis is on the decision-focused learning (DFL) framework, also recognized as "predict-then-optimize," which seamlessly integrates prediction and optimization into a unified end-to-end system, streamlining the data-decision pipeline. More specifically, we introduce a DFL framework named \emph{G-DFL4CO}, employing GNNs to address COPs. Addressing these notoriously challenging NP-hard problems, we present two interconnected modules: (1) a predictive model based on a graph neural network trained in an unsupervised or self-supervised manner, and (2) an optimizer utilizing a quadratic binary unconstrained optimization solver. We bridge the gap between neural networks and combinatorial optimization. To showcase the efficiency and precision of our framework, we conduct empirical evaluations on several well-established COPs, including maximum cut, maximum independent set, and minimum vertex cover.


Our contributions are summarized as follows:

\begin{itemize}
\item Our strategy for combinatorial optimization revolves around adopting a decision-making perspective and crafting a decision-focused framework that seamlessly integrates prediction and optimization into a unified end-to-end system. Through the utilization of this framework, our objective is to enhance the efficiency and precision of combinatorial optimization, offering a more effective solution to intricate optimization problems.

\item For addressing combinatorial optimization problems with GNNs, we introduce a decision-focused learning framework named \textbf{G-DFL4CO}. This framework comprises two meticulously crafted modules: a graph predictive model trained using unsupervised/self-supervised learning techniques and a quadratic binary unconstrained optimization solver. The integration of these modules offers a robust and efficient methodology for combinatorial optimization, capable of handling complex problems and delivering precise results.

\item In our investigation, we conduct a series of experiments on renowned COPs, including the maximum cut, maximum independent set, and minimum vertex cover. The outcomes of these experiments allow us to showcase that our proposed framework adeptly harnesses the capabilities of GNNs, delivering superior results in solving COPs when contrasted with alternative approaches.

\end{itemize}

Our work aims to address COPs and is similar to PI-GNN \cite{schuetz2022combinatorial}. Specifically, we focus on developing a decision-focused learning framework for solving such problems, which combines prediction and optimization into an end-to-end system. Our framework is inspired by the traditional decision-focused learning approach \cite{wilder2019melding}.


{\bf Outline.} The structure of our paper is organized as follows: Section~\ref{section:RelatedWorks} discusses the related works in the field, including GNNs for combinatorial optimization and DFL. Section~\ref{section:Background} presents the preliminaries required for our proposed framework. Our proposed framework is presented in Section~\ref{section:settings}, which consists of two modules that are elaborated on in Section~\ref{section:Methods1} and Section~\ref{section:Methods2}. We conduct numerical experiments and analyze the results in Section~\ref{section:Experiments}. Finally, we conclude and discuss the findings of our work in Section~\ref{section:Conclusions}.

\section{Related works}\label{section:RelatedWorks}

In this section, we present relevant literature in the domain of machine learning for combinatorial optimization (ML4CO), encompassing topics such as graph neural networks, combinatorial optimization, and decision-focused learning. The latter is a framework that facilitates learning differentiable optimizers with a focus on discrete optimization. Additionally, we elucidate various perspectives and connections between these works and our contributions.

\paragraph{Graph neural networks} In the realm of graph neural networks (GNNs), diverse models have been developed, such as graph convolutional networks (GCNs) \cite{kipf2016semi}, graph attention networks (GATs) \cite{velickovic2018graph}, graph isomorphism networks (GINs) \cite{xu2018how}, and graph transformer networks \cite{yun2019GTN,yun2022FastGTN}. These models have showcased robust performance across a spectrum of graph-based tasks. GCNs employ a localized first-order approximation of spectral graph convolutions, facilitating effective scalability to large graphs. Conversely, GATs employ an attention mechanism to evaluate the influence of different neighbors, accommodating sparse and irregular graphs. GINs, rooted in a permutation-invariant function that aggregates neighboring node features, demonstrate resilience to graph isomorphism. Lastly, graph transformers apply the transformer architecture \cite{yun2019GTN,yun2022FastGTN} to graphs, capturing higher-order interactions between nodes.

\begin{figure}[t]
    \centering
    \includegraphics[width=1.0\linewidth]{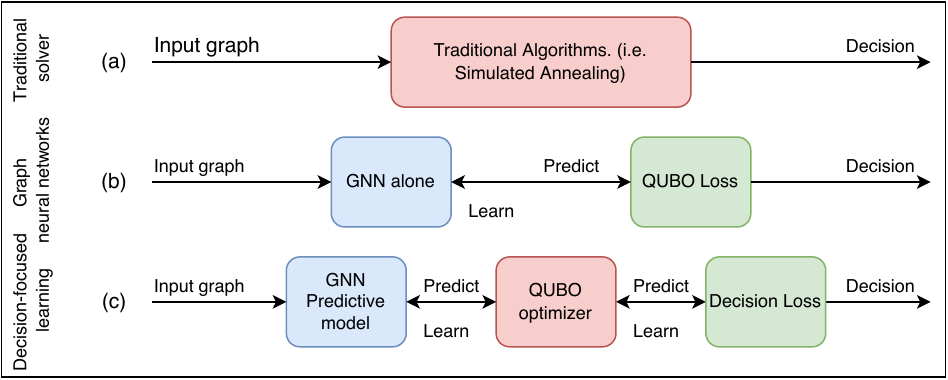}
    \caption{The overarching framework of our proposed G-DFL4CO involves an end-to-end decision-focused learning process.}
    \label{fig: G-DFL4CO}
\end{figure}



\paragraph{Combinatorial optimization} Combinatorial optimization, a field of extensive research \cite{boettcher2022inability,angelini2022cracking,schuetz2022combinatorial,liu2021decision}, is dedicated to discovering optimal solutions for complex problems characterized by numerous constraints and variables. Various studies have delved into different facets of combinatorial optimization problems and solution techniques. \cite{kochenberger2006unified} offers an overview of major combinatorial optimization problems, including the traveling salesman problem and job market scheduling, discussing both exact and heuristic solution methods for these NP-hard problems. Recent surveys have honed in on specific techniques like local search methods \cite{grasas2016simils}, linear programming algorithms \cite{raidl2008combining}, and constraint programming \cite{rossi2008constraint}. By framing problems using graphs and employing combinatorial optimization techniques, researchers can devise efficient algorithms to address real-world problems. This research area finds applications in various fields such as transportation \cite{triki2014stochastic}, telecommunications \cite{resende2003combinatorial}, manufacturing \cite{crama1997combinatorial}, and logistics \cite{sbihi2010combinatorial}, among others. Consequently, combinatorial optimization on graphs assumes a pivotal role in solving intricate problems and augmenting the world's efficiency and functionality.

\paragraph{Decision-focused learning} Decision-focused learning is an instructional approach that prioritizes the acquisition of knowledge and skills directly applicable to improving decision-making. This approach holds particular relevance in fields where decision-making is pivotal, such as business, finance, and healthcare. It involves end-to-end learning with a two-stage task using gradients in domains where optimization layers are available \cite{wilder2019melding,shah2022decision,wang2020automatically}. In today's fast-paced, data-driven world, decision-focused learning is gaining popularity, recognizing the crucial role of making informed decisions swiftly and accurately for success. By arming learners with the necessary tools and knowledge for improved decision-making, decision-focused learning contributes to organizations achieving their objectives and maintaining a competitive edge. The primary focus of this paper is to further advance and refine the existing framework, with the ultimate goal of applying it to the field of combinatorial optimization. Combinatorial optimization involves determining the best possible solution from a multitude of combinations and permutations, often in complex and dynamic environments.

\begin{figure*}[t]
    \centering
    \includegraphics[width=0.8\linewidth]{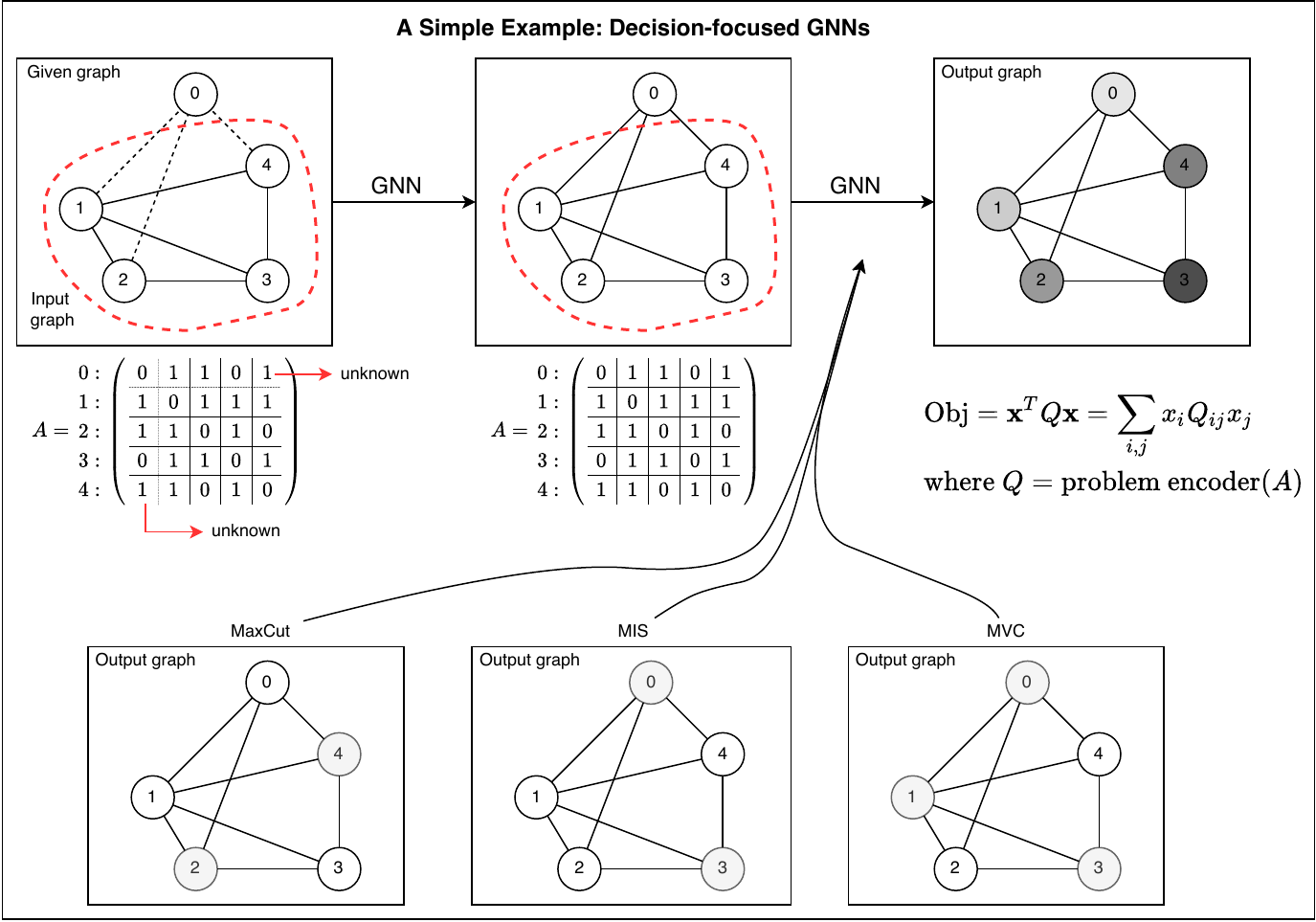}
    \caption{Flow chart illustrating the end-to-end workflow for our proposed framework using a simple example with five nodes.}
    \label{fig: example}
    \vspace{10pt}
\end{figure*}

\section{Mathematical background}\label{section:Background}

\paragraph{Notations.} We denote $\mathcal{G} = (\mathcal{V}, \mathcal{E})$ for the input graph with $N$ nodes $\{{v}_i\}_{i=1}^N$ and $M$ edges $\{e_i\}_{i=1}^M$, where $e_i = (v_{i_1}, v_{i_2}) \subset \mathcal{V} \times \mathcal{V}$. The sum of the edge weights related to node $v$ is denoted by $d(v)$, and $\mathbf{d} = \{d(v)\}_{v \in \mathcal{V}} \in \mathbb{R}^N$ gets the node degrees.  Also, we define matrix $\mathbf{D}$ as the degree matrix whose diagonal elements are obtained from $\mathbf{d}$. An overall of notations are presented in Table~\ref{table:notation}.

\subsection{Graph and graph neural networks}

Given a graph $\mathcal{G} = (\mathcal{V}, \mathcal{E}, \mathcal{A}, \mathcal{X})$, where $\mathcal{V} = \{v_i\}_{i=1}^{N} \in \mathbb{R}^N$, $\mathcal{E} = \{ (v_{i_1},v_{i_2}) \}_{i = 1}^m \subset \mathcal{V} \times \mathcal{V}$ denote the set of nodes and edges, respectively. $\mathcal{A} \in \mathbb{R}^{N \times N}$ denotes the adjacency matrix of $\mathcal{G}$ and $\mathcal{X} = [x_1, x_2, \cdots, x_N] \in \mathbb{R}^{N \times D}$ represents the feature space of nodes and $D$ is the dimensional of node feature. $\mathcal{A}$ describes the connection status for all the node pairs. If $\mathcal{A}(i,j)=1$ means node $i$ and $j$ are connected, else $\mathcal{A}(i,j)=0$, otherwise.

\begin{table}[t]
    \caption{ \small{Notations} } \label{table:notation}
    \centering
    \begin{tabular}{lp{4cm}}
     \toprule
        Notations & Descriptions  
     \\ \midrule
        $\mathcal{G}, \mathcal{G}^{\prime} \subseteq \mathcal{G}$ & given (input) graph  \\
        $\mathcal{V} = \{v_i\}_{i=1}^N, \mathcal{V}^{\prime} \subseteq \mathcal{V}$ & set of nodes \\
        $\mathcal{E} = \{e_i\}_{i=1}^M, \mathcal{E}^{\prime} \subseteq \mathcal{E}$ & set of edges \\
        $\mathcal{A} \in \{0,1\}^{N \times N}, \mathcal{A}^{\prime} \subseteq \mathcal{A}$ & adjacency matrix
        \\ 
        \midrule
        $\mathbf{H}(Q)$ & QUBO based Hamiltonian \\
        $f$ & objective function \\
        $\mathbf{x}$ & decision variable 
     \\ \bottomrule
    \end{tabular}
    \vspace{10pt}
    \end{table}

\begin{table}[t]
    \caption{  Problem setup and their objective functions. } \label{table: Problem}
    \centering
         \begin{tabular}{ll} 
        \toprule
        Task 
        & Problem (optimization form) ($x_i \in \{0,1\}$)
        \\
        \midrule
        MaxCut 
        & $\max\sum_{(i,j)\in E} x_i+x_j-2x_ix_j$
        \\
        MIS 
        & $\max \sum_{i\in V}x_i \quad \text{s.t.} \,  x_i+x_j \leq 1,\ \forall  (i,j)\in E$ \\
        MVC 
        & $\min \sum_{i\in V}x_i \quad \text{s.t.} \, x_i+x_j\geq 1 ,\  \ \forall (i,j)\in E$ \\
        \bottomrule
         \end{tabular}
         \vspace{10pt}
    \end{table}

Given that the combinatorial optimization in our work can be regarded as a node classification task, we introduce this task first. 
Node classification involves assigning categories or labels to nodes in a network. In this context, nodes symbolize entities in the network and edges represent relationships between nodes. 
The process of classifying the nodes in a network aids in identifying nodes with similar connections or properties. 
In a general problem setting, the labels of $N$ nodes are given by $Y = [y_1, y_2, \cdots, y_N] \in \mathbb{R}^{N \times L}$, 
where $L$ denotes the number of categories and $y_i$ is a soft one-hot vector $\sum_{j} y_{ij}=1$. The final decision for model prediction is $\hat{y} = \arg \max_j y_{ij}$. 
The target is to learn a classifier from the labeled nodes, formally,

\vspace{-0.5cm}
\begin{align*}
    \operatorname{GNN}(\mathcal{A}, \mathcal{X} | \theta) = f(\mathcal{N}({x}), {x}| \theta),
\end{align*}
where $\theta$ denotes the parameters of the classifier and $\mathcal{N}(x)$ denotes the neighbors of $x$.

In this part, we introduce the GNNs in detail. Without loss of generality, we give a brief introduction to GCNs. It is originally proposed by \cite{kipf2016semi}. The graph layer can be explicitly expressed as follows:

\vspace{-0.5cm}
\begin{align}\label{equ: GNNlayer}
    \mathcal{H}^{k+1} = \sigma (\hat{\mathcal{A}} \mathcal{H}^k \mathcal{W}^k),
\end{align}
where $\mathcal{H}^k = [h_1^k, h_2^k, \cdots, h_N^k]$ is the $k$-th layer of GCNs and $h_i^k$ is the hidden vector for node $i$ ($i=1,2,\cdots,N$) and $\hat{\mathcal{A}} = \hat{\mathcal{D}}^{-\frac{1}{2}}(\mathcal{A}+\mathcal{I})\hat{\mathcal{D}}^{-\frac{1}{2}}$ is the re-normalization of the adjacency matrix, 
where $\hat{\mathcal{D}}$ is the corresponding degree matrix of $\mathcal{A}+\mathcal{I}$. $\sigma(\cdot)$ is the activate function (i.e. ReLU, tanh). We denote the mapping computed by Equation~\ref{equ: GNNlayer} as one layer GCN in the following sections.

\subsection{Combinatorial optimization}\label{section: defitions}

This section introduces several canonical combinatorial optimization problems, such as the maximum cut problem (MaxCut), maximum independent set problem (MIS), and minimum vertex cover problem (MVC), among others.

\begin{definition}[Maximum cut (MaxCut)]\label{def: maxcut}
    Given a graph $\mathcal{G} = (\mathcal{V}, \mathcal{E})$, a maximum cut, denoted as $\mathcal{V}^*$, is a cut whose size equals or surpasses that of any other cut. That is, it is a partition of the set $V$ into two sets $V^*$ and $\mathcal{V}^{*C} = \mathcal{V}/{\mathcal{V}^*}$, such that the number of edges between $\mathcal{V}^*$ and $\mathcal{V}^{*C}$ exceeds that of any other partition. The task of finding such a partition is referred to as the max cut problem (MaxCut).
\end{definition}

MaxCut finds applications in diverse fields, including physics \cite{barahona1988application}, social network analysis \cite{kochenberger2013solving}, and image segmentation \cite{de2013estimation,dunning2018works}. The resolution of MaxCut poses a challenging task, necessitating the creation of efficient algorithms and heuristics.


\begin{definition}[Maximum independent set (MIS)] \label{def: mis}
    Given an undirected graph $\mathcal{G}=(\mathcal{V}, \mathcal{E})$, an independent set is defined such that any two points in the set are not connected by edges. The task of identifying the largest independent set in the graph $\mathcal{G}$ is referred to as the maximum independent set problem (MIS).
\end{definition}

The issue at hand holds substantial applications across various domains, including wireless network design \cite{park2017wireless}, social network analysis \cite{chuang2022social}, and computational biology \cite{samaga2010computing}. The identification of the maximum independent set poses an NP-hard problem, signifying the absence of any known efficient algorithm capable of universal resolution. Consequently, researchers have devised a range of approximation algorithms and heuristics to seek approximate solutions for this problem.


\begin{definition}[Minimum vertex cover (MVC)] \label{def: mvc}
     Given an undirected graph $\mathcal{G}=(\mathcal{V}, \mathcal{E})$, a vertex cover set is defined such that each edge in the graph $\mathcal{G}$ has at least one of its endpoints included in the set. The task of identifying the vertex covering set with the fewest number of vertices in the subset is referred to as the minimum vertex cover problem (MVC).
\end{definition}

The issue at hand holds significant applications across diverse fields, including network design \cite{zhang2014design}, scheduling \cite{choi2011area}, and facility location \cite{holmberg2001experiments,li2021distributed}. The identification of the minimum vertex cover represents an NP-hard problem, indicating the absence of any known efficient algorithm capable of solving it universally. Consequently, researchers have devised a spectrum of approximation algorithms and heuristics to discover approximate solutions for this problem.



\section{Our proposed framework}\label{section:settings}

We consider an inductive setting for our problem which includes both learning and optimization. Our input is the subset of the graph, while the testing set is the whole graph. The input graph $\mathcal{G}^{\prime} = (\mathcal{V}^{\prime}, \mathcal{E}^{\prime})$ is somehow partially observed and we will perform the combinatorial tasks on the whole graph $\mathcal{G} = (\mathcal{V}, \mathcal{E})$, where $\mathcal{E}^{\prime} \subset \mathcal{E}, \mathcal{V}^{\prime} \subset \mathcal{V}$ ($\mathcal{G}^{\prime}$ is a sub-graph of $\mathcal{G}$). This setting enables our model to be highly extensible and adaptable, even to unsupervised learning.

Consider $\mathcal{A}^{\prime}$ and $\mathcal{A}$ as the adjacency matrices in the training set and the original matrix, respectively. The learning task aims to derive the $\mathcal{A}$ from $\mathcal{A}^{\prime}$. For the objective function, we have to introduce a decision variable $\mathbf{x} \in \{0,1\}^{|\mathcal{V}|}$ for the nodes $|\mathcal{V}|$ and the optimization problem is presented as follows,

\vspace{-0.5cm}
\begin{align}\label{equation: setup}
    \min_{\mathbf{x}} f(\mathbf{x}, \mathcal{A}),
\end{align}
where $f$ is responding to the specific problem.

Prior to the optimization stage for $\mathbf{x}$, it is necessary to learn $\mathcal{A}$ from $A^{\prime}$, denoted as $\mathcal{A} = \Phi(\mathcal{A}^{\prime}, \xi)$. Consequently, we formulate an end-to-end optimization problem spanning from the input $\mathcal{A}^{\prime}$ to the decision $\mathbf{x}$, as illustrated below.

\vspace{-0.5cm}
\begin{align}\label{equation: setup_}
    \min_{\mathbf{x} \in \mathcal{F}} f(\mathbf{x}, \mathbb{E}_{(\mathcal{A}, \xi) \sim \mathcal{D}} [\Phi(\mathcal{A}^{\prime}, \xi)]),
\end{align}

To enhance comprehension of our framework, an illustrative depiction is provided in Figure~\ref{fig: G-DFL4CO}. The framework shares similarities with traditional decision-focused learning, integrating two primary modules: learning and decision-making. Expanding on these two modules, discussed in Section~\ref{section:Methods1} and Section~\ref{section:Methods2}, we will provide a comprehensive and accessible explanation of our framework, G-DFL4CO. This entails a comparison with other baseline approaches within the overarching framework depicted in Figure~\ref{fig: G-DFL4CO}. Subsequent sections will delve into the details of each module.


\noindent{\bf Decision-focused Graph Neural Networks.}\; It is an end-to-end framework, that represents input graphs or decisions on the graph by mapping them to a decision space. The learning problem can be formulated as equation \ref{equation: setup_}.

\subsection{Graph predictive model}\label{section:Methods1}

The learning module utilized in our framework is based on Graph Neural Networks (GNNs), specifically employing the original GCNs introduced by \cite{kipf2016semi}, and consists of two layers. The task formulated in Equations~\ref{equation: setup} and \ref{equation: setup_} involves predicting the entire graph $\mathcal{G}$ using the graph model, with the subgraph $\mathcal{G}^{\prime}$ as the input. This enables the execution of the downstream task in an inductive setting. In our experiments, we maintained the accuracy of the framework while sampling only 80\% of the nodes in $\mathcal{G}$. Referring to the first image in Figure~\ref{fig: example}, it is observed that node 0 remains unobserved. Therefore, for reconstructing the adjacency matrix $\mathcal{A}$ of the entire graph $\mathcal{G}$, it is essential to input the graph to predict the connection status of node $0$.

\subsection{Optimizer}\label{section:Methods2}


In this section, we introduce the GNN optimizer, which is inspired by PI-GNN \cite{schuetz2022combinatorial}. To frame our problem, we utilize quadratic binary unconstrained optimization (QUBO), and we use a differentiable loss function to train our framework.

Following the existing work in discrete optimization \cite{schuetz2022combinatorial}, QUBO can be expressed as the following form from Hamiltonian:

\vspace{-0.2cm}
    \begin{align}
        \mathbf{H}_{QUBO} = \mathbf{x}^{\mathrm{T}} Q\mathbf{x} = \sum_{ij} \mathbf{x}_i Q_{ij} \mathbf{x}_j,
    \end{align}
where $\mathbf{x} = (\mathbf{x}_1,\mathbf{x}_2,\cdots,\mathbf{x}_n)$ is a vector of binary variables for the node decision and the matrix $Q$ is a square matrix of constant numbers, tailored to the actual problem to solve.

We consider a maximum problem in a discrete space and denote the function as $f: \{0,1\}^{N \times N} \longrightarrow \mathbb{R}$. We follow the work \cite{chen2020measuring} and confine our attention to submodular functions that are monotone (i.e. $f(\mathcal{G} \cup \{v\}) - f(\mathcal{G}) \geq 0$) and normalized (i.e. $f(\emptyset) = 0$). Through these settings, our framework can easily accommodate more general constraints.

After we compute the objective value defined in Section~\ref{section: defitions}, we can obtain an output of our framework for updating the parameters. We can then differentiate the traditional solver to build an end-to-end framework and update parameters through the solvers. More specifically, the problem can be transformed into an unconstrained one with penalty terms and use multilinear extensions $F(\mathbf{x}, \theta)$ as the objective function instead. Then the gradient can be computed as $\frac{d}{d\mathbf{x}}F(\mathbf{x}, \theta)$. When the parameters $\theta$ are known, the gradient can be expressed in the following form:

$$
    \frac{d}{d \theta_{k j}} \nabla_{\mathbf{x}_{i}} F(\mathbf{x}, \theta)=\left\{\begin{array}{ll}
-\theta_{i j} \mathbf{x}_{k} \prod_{\ell \neq i, k} 1-\mathbf{x}_{\ell} \theta_{\ell j} & \text { if } k \neq i, \\
\prod_{k \neq i} 1-\mathbf{x}_{k} \theta_{k j} & \text { otherwise. }
\end{array}\right.
$$

Above all, we employ a differentiable loss function as follows to train our framework:

\vspace{-0.5cm}
\begin{align}
    \mathbf{L} = \mathbf{H}_{QUBO} + \lambda \mathbf{L}_{obj},
\end{align}
where $\lambda$ is a coefficient to balance the importance between GNNs and traditional solvers.

\begin{definition}[local maxima]
    We say $\mathbf{x}^*$ is the local maxima of $F(\mathbf{x}, \theta)$ if it satisfies the following conditions:
    \begin{itemize}
        \item $\nabla_{\mathbf{x}} F(\mathbf{x}^*, \theta) = 0$;
        \item $\nabla^2_{\mathbf{x}} F(\mathbf{x}^*, \theta) \succ 0$.
    \end{itemize}
\end{definition}

\begin{theorem}
    Assume $\mathbf{x}^*$ is the local maxima of $F(\mathbf{x}, \theta)$, there exists a neighborhood $\mathcal{I}$ around $\mathbf{x}^*$ such that the maximizer of $F(\mathbf{x}, \theta)$ within $\mathcal{I} \cup \mathcal{X}$ is differentiable almost everywhere.
\end{theorem}

\begin{figure}[t]
    \centering
    \includegraphics[width=1.0\linewidth]{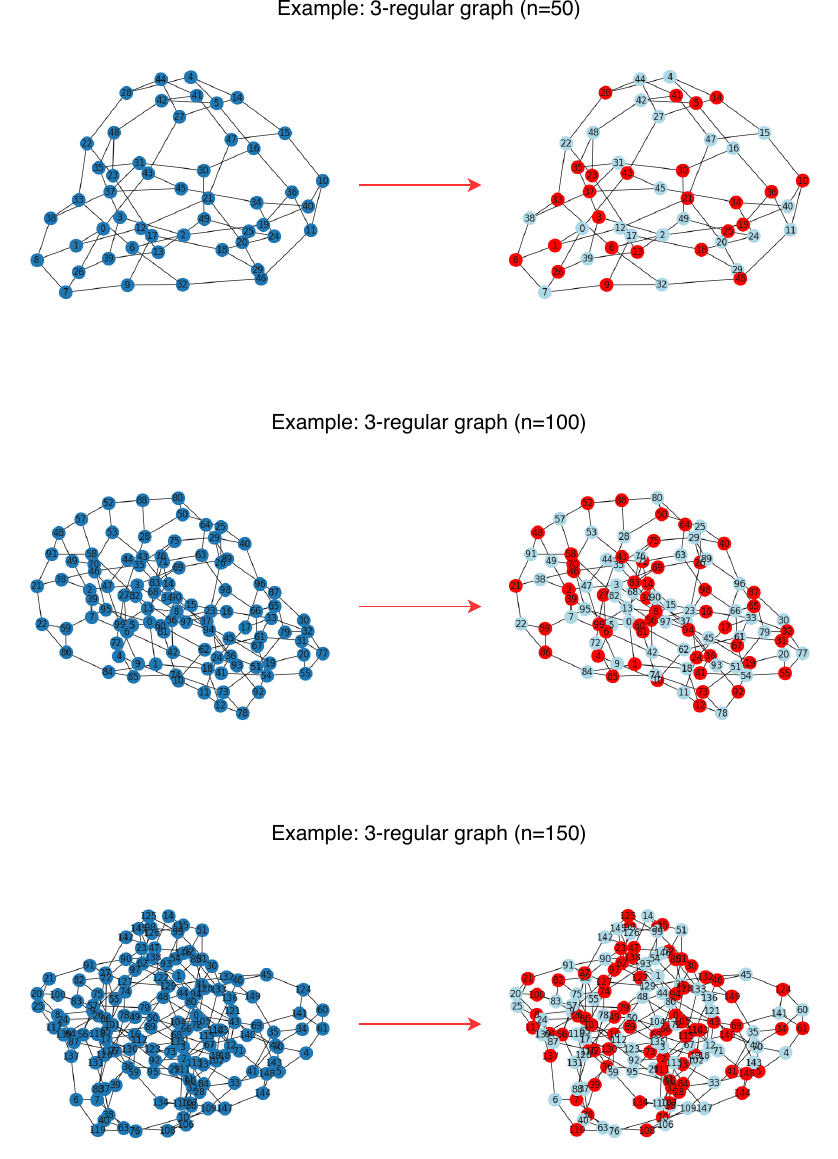}
    \caption{Some visualization examples of the d-regular instances ($n=50,100,150$).}
    \label{fig: maxcut2}
\end{figure}

\section{Experiments}\label{section:Experiments}

\begin{figure*}[t]
    \centering
    \includegraphics[width=0.8\linewidth]{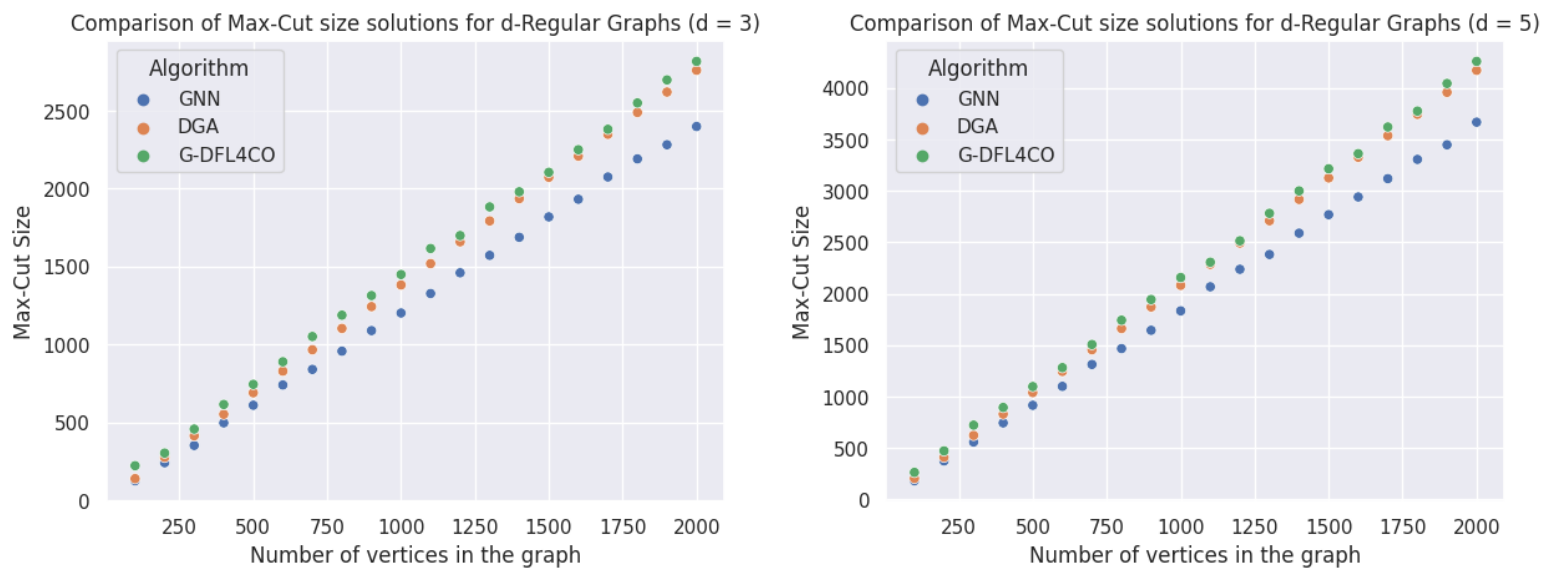}
    \caption{Comparison of MaxCut solution for d-regular graphs ($d = 3,5$) in different size graphs.}
    \label{fig: maxcut}
\end{figure*}

\begin{figure*}[t]
    \centering
    \includegraphics[width=\linewidth]{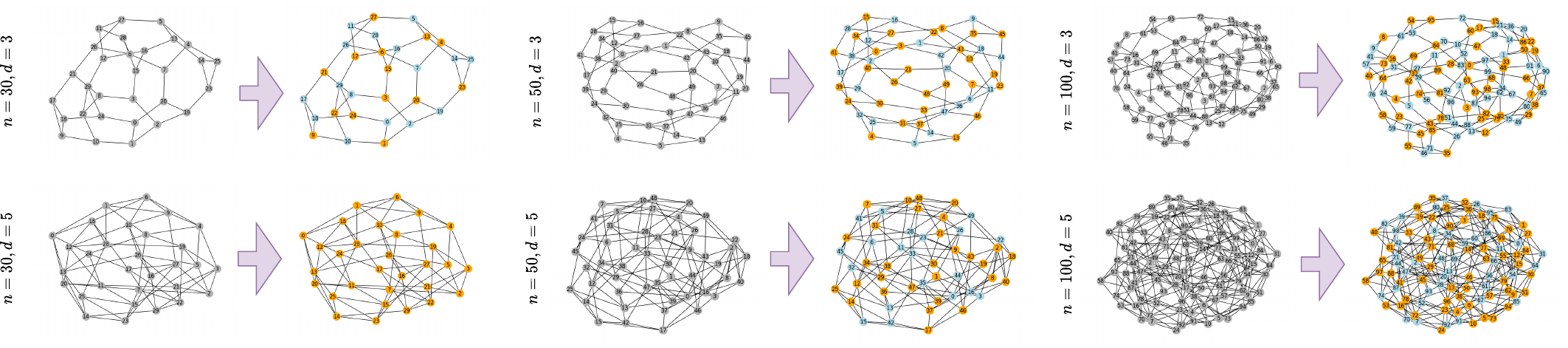}
    \caption{Some visualization of Minimum Vertex Cover for several d-regular graphs in the framework of G-DFL4CO.}
    \label{fig: mvc}
\end{figure*}

\begin{figure*}[t]
    \centering
    \includegraphics[width=1\linewidth]{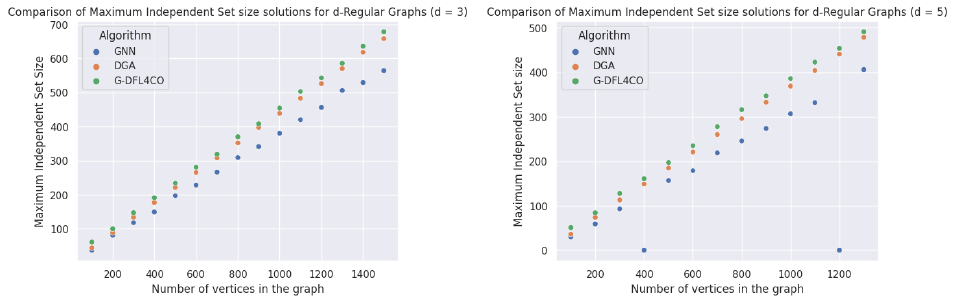}
    \caption{Experimental results for Maximum Independent Set for d-regular graphs in different sizes.}
    \label{fig: mis}
\end{figure*}

In this section, we conduct experiments to demonstrate the effectiveness of our G-DFL4CO framework on existing combinatorial optimization using benchmark datasets and manually constructed regular graphs. First, we introduce the experimental setup in Section 5.1 including the description of each evaluation task and baseline methods for the three scenarios --- MaxCut, MIS, and MVC. In Section 5.2, we present overall experimental results in Table~\ref{table: maxcut}, Figure~\ref{fig: maxcut}, Figure~\ref{fig: mis}, and Figure~\ref{fig: mvc}. We compare the results and running time with a degree-based greedy algorithm (DGA) and a GNN-based model (GNN). In Section 5.2, we also give a discussion of the framework and experimental results. We release our code in \url{https://anonymous.4open.science/r/GDFL-79A6/}.

\begin{table*}[t]
\small
    \caption{  Experimental results for MaxCut on some Gset instances. $\epsilon$ denotes the relative error. The metric is the size of Cut and larger is better. } \label{table: maxcut}
    \centering
         \begin{tabular}{cccccccccc} 
        \toprule
        \multirow{2}{*}{\textbf{ Graph }}
        & \multirow{2}{*}{\textbf{ nodes }}
        & \multirow{2}{*}{\textbf{ edges }}
        & \multicolumn{6}{c}{\textbf{ Methods }} 
        & \multirow{2}{*}{\textbf{ $\epsilon$ }}\\
        \cline{4-9}
        &&& BLS 
        & DSDP 
        & KHLWG 
        & RUN-CSP 
        & PI-GNN 
        & \textbf{G-DFL4CO}\\
        \midrule
        G14 & 800 & 4694 & \textbf{3064} & 2922 & 3061 & 2943 & 3026 & \textbf{3060} & 0.13\%\\
G15 & 800 & 4661 & \textbf{3050} & 2938 & \textbf{3050} & 2928 & 2990 & {3038} & 0.39\%\\
G22 & 2000 & 19990 & \textbf{13359} & 12960 & \textbf{13359} & 13028 & 13181 & 13333 & 0.19\%\\
G49 & 3000 & 6000 & \textbf{6000} & \textbf{6000} & \textbf{6000} & \textbf{6000} & 5918 & \textbf{6000} & 0\%\\
G50 & 3000 & 6000 & \textbf{5880} & \textbf{5880} & \textbf{5880} & \textbf{5880} & 5820 & 5860 &0.34\%\\
G55 & 5000 & 12468 & \textbf{10294} & 9960 & \textbf{10236} & 10116 & 10138 & 10162 & 1.28\%\\
G70 & 10000 & 9999 & \textbf{9541} & 9456 & 9458 & — & 9421 & \textbf{9499} & 0.44\%\\
        \bottomrule
         \end{tabular}
    \end{table*}

\subsection{Scenarios}

First, we briefly introduce the three scenarios used in our experiments.

{\bf 1).} MaxCut is a well-known optimization problem in computer science and mathematics. The problem involves dividing a given set of objects into two subsets such that the sum of weights of the edges between the two subsets is maximized. In essence, it seeks to find a cut that separates the two subsets in a way that maximizes the total weight of the edges crossing the cut. The definition of this problem has been previously described in Definition~\ref{def: maxcut}.

{\bf 2).} MIS is a fundamental problem in graph theory and computer science. Given a graph, an independent set is a set of vertices in which no two vertices are adjacent. The MIS problem seeks to find the largest possible independent set in a given graph. This problem has been previously defined in Definition~\ref{def: mis}.

{\bf 3).} MVC is a renowned problem in graph theory and computer science. Given a graph, a vertex cover is a set of vertices that covers all the edges in the graph. The MVC problem seeks to find the smallest possible vertex cover in a given graph. This problem has been previously defined in Definition~\ref{def: mvc}.


\paragraph{Datasets.} (1) We have conducted supplementary experiments on Max-Cut benchmark instances and their work on random d-regular graphs. These experiments were performed using the publicly available Gset dataset\footnote{https://web.stanford.edu/~yyye/yyye/Gset/}, which is commonly used for testing Max-Cut algorithms. The results of these experiments are publicly available. The purpose of conducting these experiments was to provide a more comprehensive analysis of the performance of Max-Cut algorithms, as the Gset dataset contains a diverse set of instances that are representative of real-world problems. By testing their algorithm on these instances, we evaluated its performance under a range of conditions and determined its effectiveness in solving Max-Cut problems. (2) Regular graphs.

\paragraph{Baselines.} We compare with some competitive methods: physics-inspired GNN solver (PI-GNN) \cite{schuetz2022combinatorial}, an SDP solver using dual scaling (DSDP) \cite{DBLP:journals/na/LingX12},  a combination of local search and adaptive perturbation referred to as Breakout Local Search (BLS) \cite{DBLP:journals/eaai/BenlicH13} which provides the best-known solutions for the Gset data set, a Tabu Search metaheuristic (KHLWG) \cite{DBLP:journals/heuristics/KochenbergerHLWG13} and a recurrent GNN architecture for maximum constraint satisfaction problems (RUN-CSP) \cite{DBLP:journals/corr/abs-1909-08387}.

\subsection{Results and discussions}

We provide testing results for MaxCut on some Gset instances in Table~\ref{table: maxcut}. G-DFL4CO outperforms PI-GNN across all the instances and remains relatively error below 1\%. Additionally, we test our framework on the regular graphs in different sizes for the three tasks. Our experimental results surpassed those of DGA and standalone GNN, while also demonstrating shorter run times. The results can be seen in Figure~\ref{fig: maxcut}, Figure~\ref{fig: mis}, and Figure~\ref{fig: mvc}.


We assert that our framework is effective in solving three significant COPs. We also highlight the importance of incorporating DFL into the process of solving such problems. According to the above results, this approach can lead to more efficient solutions. Additionally, we suggest that this sight has identified a potential link between GNNs and traditional algorithms used for NP problems. Overall, the experimental results support the effectiveness of their framework in addressing COPs, and we believe that these findings could have significant implications for future research in this field. 

In DFL framework, we can consider different exploration strategies to generate new sets of edges and use the loss function to evaluate the performance of these edge sets. In addition to randomly selecting some edges and cutting them, greedy algorithms or other heuristic algorithms can also be used to generate new edge sets in the MaxCut problem. These algorithms can often produce high-quality edge sets in a short time, but they may get stuck in local optima. Furthermore, DFL can be combined with other optimization techniques such as genetic algorithms, simulated annealing, and local search. These techniques can help DFL to escape local optima and find better decisions. Overall, DFL is a powerful machine learning method that can be used to solve various COPs. In the MaxCut problem, DFL can guide the choice of which edges should be cut to maximize the total weight of the cut. Although MaxCut is an NP-hard problem, DFL can produce high-quality solutions in a reasonable amount of time.


\section{Discussion and conclusion}\label{section:Conclusions}



\paragraph{Discussion} Combinatorial optimization problem (e.g.  traveling salesman problem, maximum cut, and  satisfiability problem) has a wide range of beneficial applications in engineering and computer science, including stock market portfolio optimization, budget allocation, and network routing. First, G-DFL4CO may encode the bias present in the training graph, which leads to stereotyped predictions when the prediction is applied to real-world applications. Second, some harmful network activities could be augmented by powerful GNNs, e.g., spamming, phishing, and social engineering.  We expect future research should help to resolve these problems.

\paragraph{Conclusion} In this study, we have developed a framework that is centered on decision-making to tackle COPs. The framework employs unsupervised GNNs as its predictive model, and the QUBO optimizer serves as its decision-making module. Through our experiments, we have demonstrated the effectiveness of the framework in addressing three crucial COPs. Moreover, we believe that our work has paved the way for a more effective and efficient solution to NP problems by connecting the gap between GNNs and conventional algorithms.

\bibliographystyle{named}
\bibliography{ijcai24}

\end{document}